\documentclass[letterpaper, 10 pt, conference]{ieeeconf}

\IEEEoverridecommandlockouts
\usepackage{amsmath}
\usepackage{amssymb}
\usepackage{graphicx}

\let\oldthebibliography\thebibliography
\renewcommand{\thebibliography}[1]{\oldthebibliography{#1}\setlength{\itemsep}{0pt}\setlength{\parskip}{0pt}}

\title{\LARGE \bf
Beyond Task Reward: A Controller-Restriction Protocol\\
for Evaluating Embodiment-Dependent Competence
}

\author{Siyuan Zhang$^{1}$
\thanks{$^{1}$Siyuan Zhang is with the Robotics Department, University of Michigan,
        Ann Arbor, MI, USA.}%
}

\begin{document}

\maketitle
\thispagestyle{empty}
\pagestyle{empty}

\begin{abstract}
Co-design methods optimize a robot's body and controller jointly and judge the result by one
number, the task reward of the fully optimized pair. That number cannot separate morphologies
whose competence depends on the controller to very different degrees. We evaluate a morphology by
restricting its controller instead, recording the task competence it retains under an explicitly
declared, low-complexity controller family, environment, task and search budget. On three EvoGym
locomotion tasks, task reward explains only $39\%$, $33\%$ and $10\%$ of the variance in this
quantity, and geometric descriptors do not predict it under run-grouped cross-validation. The
measurement is reliable across optimizer restarts (ICC$(2,k) = 0.956$--$0.986$) but depends on the
declared family: phasing the drive by actuator index instead of position ranks the same
morphologies at Spearman $0.50$--$0.63$ and reverses reward-matched pairs. As a second search
objective the axis improved competence at matched task reward in $3$ of $5$ paired runs, short of
a pre-registered bar of $4$. Used after an ordinary reward-only search instead, to choose within
its top task-reward band, it selected a different body in all $8$ runs offering a choice, at a
cost of at most $0.10$ reward units, and in $6$ of $8$ that body also scored higher under a
held-out family. Restricted-control competence is therefore a reportable property of a co-designed
morphology, interpretable only with the controller family that defines it.
\end{abstract}

\begin{figure*}[t]
  \centering
  \includegraphics[width=\textwidth]{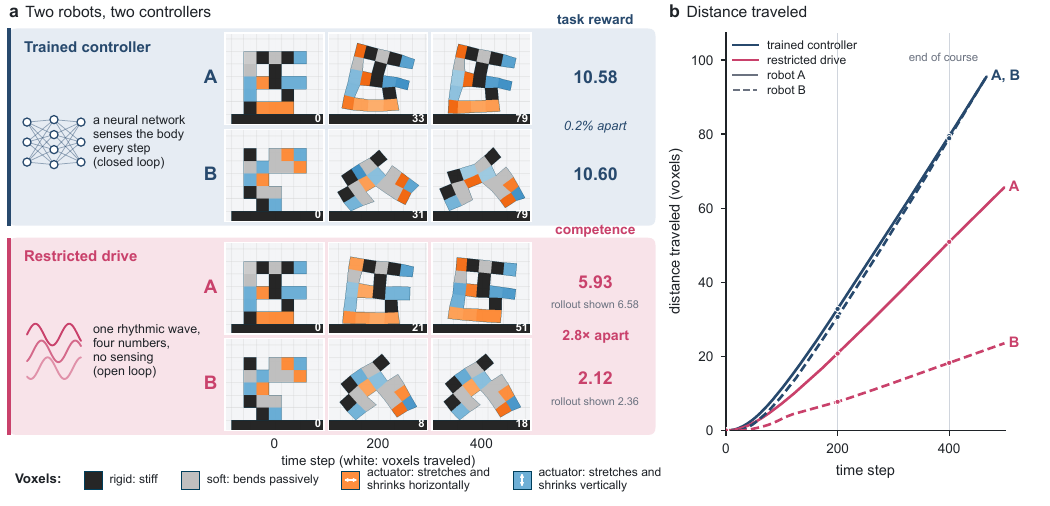}
  \caption{\textbf{Task reward does not separate robots that depend on their controller very
  differently.} Two co-designed Walker robots, A and B, built from the voxel types listed at the
  bottom. (a)~Top: each robot with the neural-network controller it was co-designed with, which
  senses the body at every step; the two reach the same task reward to within $0.2\%$ ($10.58$ and
  $10.60$). Bottom: the same bodies driven by the declared restricted family instead, one rhythmic
  wave with four parameters and no sensing; A keeps walking and B barely moves ($5.93$ against
  $2.12$, mean over three restarts; the rollout shown is the best restart). All frames share one
  window and scale, and the white number on the ground is the distance traveled in voxels.
  (b)~Distance traveled; the trained rollouts end near step $470$, at the end of the course. Under
  the ID-phased family the pair \emph{reverses} ($1.66$ against $2.06$).}
  \label{fig:problem}
\end{figure*}

\section{INTRODUCTION}

Co-design methods jointly optimize a robot's body and its controller, and are almost universally
judged by a single number: the task reward achieved by the co-optimized pair. That number says
whether the pair works, but not how much of the working depends on the controller.

Two morphologies can be indistinguishable by that number yet behave very differently once the
controller is taken away. Fig.~\ref{fig:problem} shows a pair from our runs whose fully optimized
task rewards differ by $0.2\%$, a gap no co-design report would treat as meaningful. Replacing the
trained policy with a low-complexity feedback-free drive separates them by $2.8\times$. The
distinction matters to anyone who has to build the robot, yet it is invisible in the quantity the
field publishes.

We make it measurable by restricting the controller rather than by inspecting the body: given a
morphology, we optimize within a declared family of low-complexity controllers under a declared
budget and record the task competence that remains. The declaration is the substance of the
proposal, because the score depends on the family, the task, the environment and the search
budget, and we report all four with every number. Our own data forced that emphasis. We began with
a family that phases a periodic drive by actuator index, so renumbering a fixed body's actuators
changed its score; a family phased by actuator \emph{position} removes that defect exactly. Yet
the two families rank the same $874$ morphologies at Spearman $0.50$, $0.53$ and $0.63$ and
reverse individual reward-matched pairs, including the pair in Fig.~\ref{fig:problem}. Both are
defensible, so a restricted-control result is reproducible only with its family attached, and the
useful object is a profile across families rather than a single score.

Our contributions are (i)~an evaluation protocol that quantifies the task competence a morphology
retains under a declared, permutation-invariant, low-complexity controller family, task,
environment and budget, shown to be invariant, reliable, converged and discriminative; (ii)~the
finding that task reward leaves most of this competence unexplained and geometry does not predict
it, together with the family sensitivity above; and (iii)~two uses of the measurement in design, a
pre-registered multi-objective search that falls short of its bar at a fixed generation budget for
a reason the search trajectories identify, and an exploratory screening step that changes the
chosen body at negligible task-reward cost.

\section{RELATED WORK}
\label{sec:related}

\textbf{Co-design of morphology and control.} Jointly optimizing body and controller goes back to
Sims's evolved virtual creatures~\cite{sims1994evolving} and the first physically instantiated
evolved machines~\cite{lipson2000automatic}. Evolutionary methods optimize morphology in an outer
loop and train a controller per candidate, over voxel grids~\cite{bhatia2021evolution}, graph
grammars~\cite{zhao2020robogrammar}, learned latent spaces~\cite{hu2023glso} or populations of
graph-structured agents~\cite{wang2019neural}; reinforcement-learning formulations fold the design
decision into the policy~\cite{yuan2022transform2act, gupta2021embodied, lu2025bodygen}, and a
recent line replaces the search with a generative model~\cite{wang2023diffusebot,
liu2026morphology, chen2026natural}. These methods differ in how they search but agree on what
they report: the task reward of the co-optimized pair, a joint property of a body and a controller
fitted to each other. It cannot distinguish a morphology whose competence persists under a much
simpler controller from one whose competence exists only alongside its trained policy, and nothing
in the reported quantity makes that distinction recoverable afterwards.

\textbf{Quantifying morphological computation.} That a body can perform work otherwise delegated
to a controller is a long-standing idea in embodied intelligence~\cite{pfeifer2006body},
formalized mainly through information theory as measures on the dependencies between sensors,
actions and world states~\cite{zahedi2013quantifying, ghazizahedi2015quantifying}, with a parallel
line modeling compliant bodies as physical reservoirs~\cite{hauser2011towards}. A body can take
part in that information flow while contributing nothing to the task, and the converse also holds;
our measure is behavioral instead. Closest in spirit is Haeufle \emph{et
al.}~\cite{haeufle2020morphological}, who disable a central pattern generator in a
neuromusculoskeletal model and attribute the residual oscillation to passive dynamics. The
intervention matches ours, but their read-out is an information-theoretic measure on one
biological model, not task competence across evolved morphologies.

\textbf{Restricting a controller versus measuring it.} The family we declare is a positional
traveling wave, close to the low-dimensional oscillators of central pattern generator
models~\cite{ijspeert2008central}, and serves here as an instrument: few parameters, a count
independent of body size, exact invariance to actuator relabeling, and a reimplementable
specification. Closest to this paper is Xie \emph{et al.}~\cite{xie2025morphology}, a concurrent
preprint relating morphology complexity to the computational cost of the controller driving it on
the same platform. The distinction is between \textbf{measuring} a controller and
\textbf{intervening} on it: cost in floating-point operations varies across bodies only through
the observation and action dimensionality the morphology induces, so two bodies with equal voxel
and actuator counts score the same even if an open-loop oscillator drives one and the other
collapses without feedback. Controller restriction poses that counterfactual directly. The
methodological concern is familiar from reinforcement learning, where reported performance was
shown to depend heavily on the evaluation protocol~\cite{henderson2018deep, agarwal2021precipice}.

\begin{figure*}[t]
  \centering
  \includegraphics[width=\textwidth]{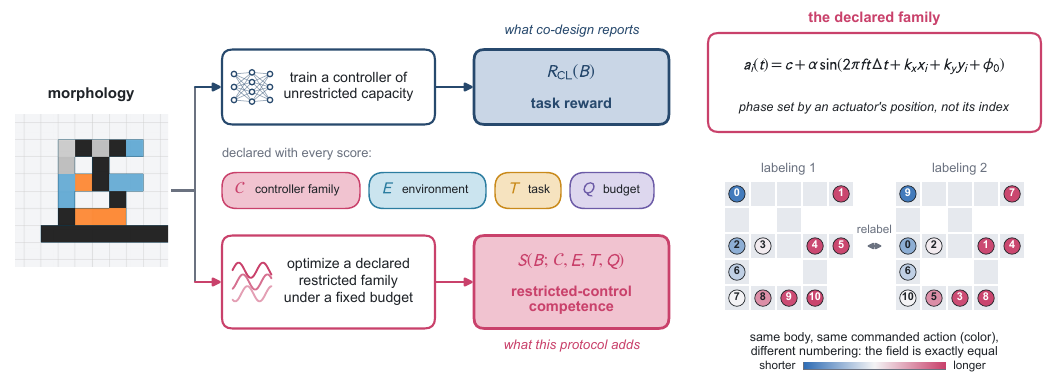}
  \caption{\textbf{The protocol, and what must be declared for the number to mean anything.}
  Upper path: what co-design reports, namely the task reward of a controller of unrestricted
  capacity. Lower path: optimize within a declared restricted family under a declared budget and
  record the competence that remains. The measured quantity is $S(B;\mathcal{C},E,T,Q)$, and each
  of the four conditioning arguments, drawn as the chips reported with every score, changes it.
  Right: phasing the drive by an actuator's \emph{position} rather than its index makes relabeling
  permute positions and actions together, leaving the action field exactly unchanged, as the inset
  shows on body A of Fig.~\ref{fig:problem}.}
  \label{fig:protocol}
\end{figure*}

\section{RESTRICTED-CONTROL COMPETENCE}
\label{sec:framework}

Write $B$ for a morphology, $E$ for the environment and its physical parameters, and $T$ for the
task and its reward. Let $\mathcal{C}$ be a family of admissible controllers, $Q$ the budget and
procedure used to search it, and $R(B, u; E, T)$ the expected return of controller $u$. The
quantity this paper defines is
\begin{equation}
S(B; \mathcal{C}, E, T, Q) \;=\; \max_{u \in \mathcal{C},\ \text{budget } Q}\ \mathbb{E}\big[R(B, u; E, T)\big].
\end{equation}
Every argument on the right changes the result, so all appear in the notation
(Fig.~\ref{fig:protocol}). A finite budget cannot return a true maximum and a best-of-$Q$ score can
only rise with $Q$, so what is computed is a budget-conditioned estimate whose convergence we
report (Sec.~\ref{sec:validity}). Two instances appear throughout: $R_{CL}(B)$ (closed loop) takes
$\mathcal{C}$ to be feedback policies of unrestricted capacity trained by reinforcement learning,
the number co-design already reports, and $R_{OL}(B)$ (open loop) takes $\mathcal{C}$ to be the
low-complexity feedback-free family below, which we call the morphology's \emph{restricted-control
competence}. Competence that survives the removal of feedback is what we informally call
morphological offloading; that phrase names an interpretation, and the definition is $S$. The pair
could be summarized as a retention ratio $R_{OL}/R_{CL}$, but its denominator is the quantity
co-design optimizes, so any intervention that raises task performance mechanically depresses it,
and on tasks whose reward crosses zero it is undefined; we therefore use the absolute $R_{OL}$ and
compare morphologies only at matched $R_{CL}$.

\subsection{A permutation-invariant controller family}
\label{sec:family}

A restricted family must be simple enough that what survives is attributable to the body and the
environment, precise enough for another group to implement, and dependent only on physically
meaningful quantities. The obvious periodic drive,
$a_i(t) = c + \alpha \sin(2\pi f t \Delta t + i\varphi)$, phases actuator $i$ by its \emph{index},
an artifact of how the body was serialized: relabeling a fixed body's actuators changes the drive
and the score. We instead phase the wave by where an actuator is:
\begin{equation}
\label{eq:family}
a_i(t) \;=\; c + \alpha \sin\!\big(2\pi f t \Delta t + k_x x_i + k_y y_i + \phi_0\big),
\end{equation}
where $(x_i, y_i)$ is the actuator's position normalized over the body's occupied bounding box,
$c$ is the actuator rest length and $\alpha$ a fixed amplitude. Relabeling permutes positions and
actions together, so the action field is unchanged \emph{exactly}. The family has four free
parameters whatever the number of actuators, so every body receives an equally expressive
controller, where a per-actuator phase vector would reward size over organization; setting
$k_x = k_y = 0$ recovers a global oscillator, so the family contains the simplest possible drive
as a strictly weaker member.

The family is a choice, and it changes the answer. On the same $874$ morphologies the ID-phased
and position-phased families agree on the ranking only moderately (Spearman $0.50$, $0.53$ and
$0.63$), and on individual reward-matched pairs they reverse: for the two morphologies of
Fig.~\ref{fig:problem}, whose task rewards differ by $0.2\%$, the position-phased family scores
$5.93$ against $2.12$ and the ID-phased family $1.66$ against $2.06$. Replacing the trained
controller also removes any body--controller interaction it had learned to exploit, so the
measured competence belongs to the pair (morphology, declared family) and is not a share
partitioned between body and controller. The quantity is \emph{the task competence accessible to a
morphology under a declared restricted controller family, task, environment and optimization
budget}, and the protocol reports one competence per declared family.

Later sections compare morphologies produced under different selection signals, so we state the
signals once. The co-design loop is held fixed and only the scalar the genetic algorithm ranks on
varies: task reward $R = R_{CL}(B)$ alone; competence $S = R_{OL}(B)$ alone; a weighted sum
$R + \lambda S$ with $\lambda = \sigma[R]/\sigma[S]$ computed rather than tuned; or Deb's
feasibility rule~\cite{deb2000efficient}, which ranks designs above a task-reward floor $\tau$ by
$S$ alone and all others by how far short they fall. Our first floor was too strict to be
informative: at $\tau = 8.24$, the $40$th percentile of the reward-only arm's late-generation task
reward on Walker, so few designs were feasible that the arm reproduced reward-only search. The
design experiment of Sec.~\ref{sec:design} instead ranks on the pair $(R, S)$ by non-dominated
sorting.

\begin{figure*}[t]
  \centering
  \includegraphics[width=\textwidth]{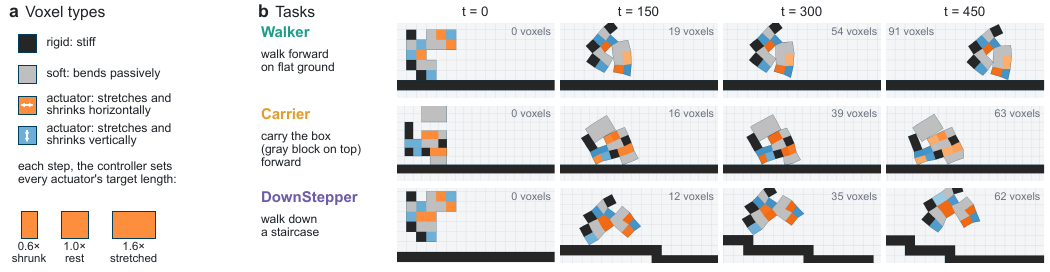}
  \caption{\textbf{The robots and the three tasks.} (a)~The four voxel types, in the simulator's
  colors. At every control step the controller sets each actuator's target length between
  $0.6\times$ and $1.6\times$ its rest length along the actuator's own axis; the resulting
  deformation against the ground is the only way a body moves. (b)~One co-designed robot per task
  under its trained controller, at four time steps, with the distance traveled so far: walk
  forward on flat ground, carry the gray box without dropping it, walk down a staircase. Each
  robot is its task's highest-reward reward-only design; on Walker, body B of
  Fig.~\ref{fig:problem}.}
  \label{fig:tasks}
\end{figure*}

\section{EXPERIMENTAL SETUP}
\label{sec:setup}

\textbf{Platform and tasks.} All experiments use EvoGym~\cite{bhatia2021evolution}, which simulates
soft robots built from voxels on a $5\times5$ grid (Fig.~\ref{fig:tasks}). Each cell is empty or
holds one of four voxel types: rigid voxels keep their shape, soft voxels deform passively, and
actuator voxels change length along one axis, horizontal or vertical, as the controller sets each
actuator's target length between $0.6$ and $1.6$ times its rest length at every control step. A
body moves only by cycling these lengths so that its deformation pushes against the ground. We use
three locomotion tasks, chosen before any measurement to span flat locomotion, object transport
and descending terrain: on Walker-v0 and DownStepper-v0 the task reward is the robot's forward
displacement, on flat ground and down a staircase, and on Carrier-v0 it averages the forward
displacement of the robot and of a box it carries, with a penalty if the box falls. A feasibility
pilot excluded Climber-v0, where the controller does not learn within our budget (a hand-designed
body reaches $0.10$, against $7.7$ on Walker), and Jumper-v0, where no random body makes forward
progress under any feedback-free drive.

\textbf{Search and training.} Morphologies come from a genetic algorithm mirroring the EvoGym
baseline: population $12$, eight generations, elitist truncation at $50\%$ and per-voxel mutation
rate $0.1$. Each design's controller is a proximal policy optimization
policy~\cite{schulman2017proximal}, trained for $600{,}000$ steps with observation normalization in
Stable-Baselines3 2.9.0~\cite{raffin2021stable} under PyTorch 2.11 and EvoGym 2.2.5; training
budget, rollout horizon ($500$ steps) and evaluation protocol are identical across every design,
arm and task. Each generation replaces half the population, so a run evaluates
$12 + 8 \times 6 = 60$ designs, and the study comprises $1{,}320$ co-designed body--controller
pairs across nine arms: every task has a reward-only and a restricted-control-driven arm, and
Walker adds one weighted and two constrained arms. Four of the six per-task arms have three
independent evolutionary runs and every other arm has two. Competence was then measured for all
$1{,}096$ distinct (task, morphology) pairs under the position-phased family, with three optimizer
restarts each; the $874$ of them in the reward-only and restricted-control-driven arms are the
population that Secs.~\ref{sec:validity} and~\ref{sec:results} analyze.

\textbf{The restricted-control measurement.} The declared family is Eq.~\eqref{eq:family} with
amplitude and rest length frozen at $\alpha = 0.5$ and $c = 1.0$, leaving $(f, k_x, k_y, \phi_0)$
free; commands below the simulator's lower bound of $0.6$ are clipped to it, which flattens the
troughs of the wave. We optimize by random search over $f \in [0.5, 2]$,
$k_x, k_y \in [-2\pi, 2\pi]$ and $\phi_0 \in [0, 2\pi)$ with budget $Q = 48$ evaluations and report
the mean of three independent restarts. Random search is used because a grid would make restart
reliability trivially perfect, and one rollout per evaluation because drive and simulator are
deterministic.

\textbf{Statistical conventions.} The independent unit is the evolutionary run: designs within a
run share a lineage, so treating $60$ of them as $60$ observations manufactures significance.
Comparisons between search objectives are computed within each run and summarized across runs, and
we report run-level effect sizes in preference to pooled $p$-values; the one design-level
exception, a set of regressions in Sec.~\ref{sec:results}, is marked where it appears. Arms are
compared only at matched task reward, and a result requires binning and regression to agree. The
retention ratio is reported but never compared on, and is undefined for $|R_{CL}| < 0.5$, which
covers $97$ of the $874$ designs. Each experiment has at most two primary endpoints, both
specified before its data existed; secondary comparisons are Holm-corrected~\cite{holm1979simple},
tests are two-sided unless marked otherwise, and non-inferiority is tested against a margin fixed
in advance.

\begin{figure*}[t]
  \centering
  \includegraphics[width=\textwidth]{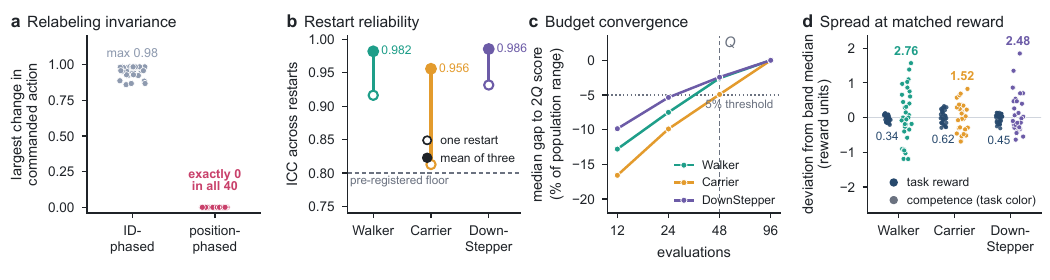}
  \caption{\textbf{The measurement is well defined}, one panel per way it could fail.
  (a) Over $40$ random relabelings of a fixed body, the ID-phased family changes the commanded
  action by up to $0.98$, while the position-phased action field is unchanged \emph{exactly}.
  (b) Optimizer restarts agree: ICC$(2,k) = 0.982$, $0.956$ and $0.986$ for the reported
  three-restart mean (filled), against $0.916$, $0.813$ and $0.932$ for a single restart (open);
  dashed, the pre-registered floor. (c) Median gap between the score at a given budget and at
  $2Q$, as a percentage of the population's score range; at $Q$ it is $2.6$, $4.9$ and $2.4\%$,
  inside the pre-registered $5\%$. (d) Within the middle decile of task reward, task reward
  spreads by only $0.34$, $0.62$ and $0.45$, competence by $2.76$, $1.52$ and $2.48$.}
  \label{fig:validity}
\end{figure*}

\section{MEASUREMENT VALIDITY}
\label{sec:validity}

A restricted-control score is a measurement only if it survives the things that have nothing to do
with the morphology. We fixed the checks below, and their pass thresholds, before running them
(Fig.~\ref{fig:validity}). \textbf{Invariance:} under the ID-phased family, relabeling a fixed
body's actuators changes the commanded action by up to $0.98$, so the drive becomes a different
drive, while under the position-phased family the action field is unchanged \emph{exactly}, as a
unit test asserts. \textbf{Reliability:} averaged over three restarts, the intraclass correlation
ICC$(2,k)$~\cite{shrout1979intraclass} is $0.982$, $0.956$ and $0.986$ on Walker, Carrier and
DownStepper against a pre-registered floor of $0.80$, and a single restart already reaches
$0.916$, $0.813$ and $0.932$. \textbf{Convergence:} a value still climbing at $Q$ would describe
the budget rather than the body, and doubling the budget raises the median design's score by
$2.6\%$, $4.9\%$ and $2.4\%$ of the population's score range, with rank agreement between $Q$ and
$2Q$ of $0.972$, $0.955$ and $0.988$, against thresholds of a median gain under $5\%$ and
$\rho \ge 0.90$. \textbf{Discrimination:} between $8\%$ and $18\%$ of morphologies sit at the
floor, so the measure is not saturated, and within the middle decile of task reward, where task
reward itself spans only $0.34$, $0.62$ and $0.45$, competence spans $2.76$, $1.52$ and $2.48$ in
the same units.

\begin{figure*}[t]
  \centering
  \includegraphics[width=\textwidth]{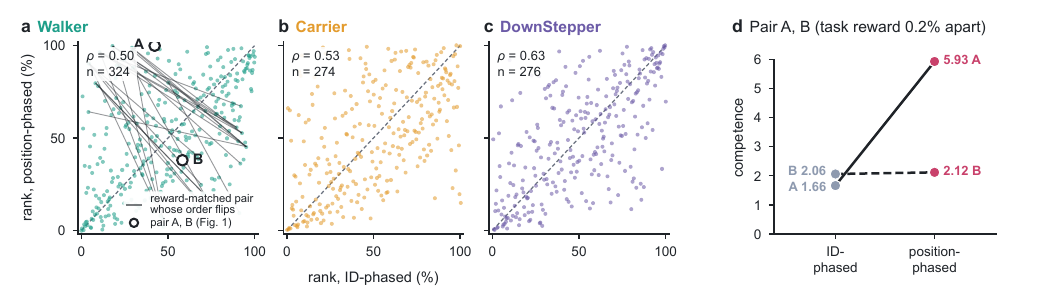}
  \caption{\textbf{Two defensible restricted families rank the same morphologies differently.}
  (a--c) Percentile rank under the position-phased family against rank under the ID-phased
  family, for every measured design; Spearman agreement is $0.50$, $0.53$ and $0.63$
  ($n = 324$, $274$, $276$). Gray segments in (a) join Walker morphologies whose task rewards are
  within $1\%$ of each other and whose ordering \emph{reverses} between the families; $649$ such
  pairs exist and the $24$ largest are drawn. Circles mark the pair of Fig.~\ref{fig:problem}.
  (d) That pair: the position-phased family separates the two bodies almost threefold, the
  ID-phased family puts them the other way round, and the lines cross.}
  \label{fig:disagreement}
\end{figure*}

\textbf{A pre-registered check that failed.} A further pre-registered check asked
whether two \emph{reasonable} position-phased families rank morphologies alike, comparing the full
wave against a variant with the vertical component removed. Agreement is $0.41$ on Walker against
a threshold of $0.70$ frozen beforehand, and $0.76$ and $0.84$ on Carrier and DownStepper. The
ID-phased and position-phased families disagree more strongly (Fig.~\ref{fig:disagreement}):
across all $874$ morphologies they agree at only $0.50$, $0.53$ and $0.63$, and they reverse
individual reward-matched pairs. Which changes of family reorder the bodies is itself informative.
A third family, C3, superposes two positional waves, the second at twice the temporal rate, mixed
so that its action range equals the declared family's exactly; it is held out for the design
experiment and never used for selection. Across the $404$ reward-only morphologies it ranks bodies
with the declared family at $0.87$, $0.87$ and $0.92$, and restricted to mixing in $[0.5, 1]$, so
that it cannot reduce to the declared family, its agreement over the $71$ screening-band designs
of Sec.~\ref{sec:design} falls by only $0.01$ and $0.06$. Adding a mode therefore leaves the
ordering largely intact, removing the vertical direction does so on Carrier and DownStepper but
not on Walker, and phasing by index does not on any task.

\textbf{A ladder of restricted families.} On the $119$ morphologies of the validity study we also
measured a constant hold (C0: $a_i(t) = L$, the held length the only free parameter) and a global
oscillator (C1: Eq.~\eqref{eq:family} with $k_x = k_y = 0$); $142$ of the $34{,}272$ ladder
evaluations ended early when the voxel mesh went unstable, and are scored up to termination rather
than excluded. Median competence is $0.06$, $-0.03$
and $0.05$ for C0, $1.26$, $0.14$ and $1.11$ for C1 and $2.96$, $1.08$ and $1.80$ for the declared
family C2, against $6.81$, $1.06$ and $3.62$ for the trained policy. C0 is a floor that says
nothing about the body, its rank correlation with C2 being $-0.11$, $+0.03$ and $+0.03$, so C2
reflects a capacity for a coordinated gait rather than resting geometry; adding the phase gradient
to a global oscillator is worth $+1.56$, $+0.70$ and $+0.79$ in the median. C1 and C2 agree at only
$+0.38$, $+0.41$ and $+0.84$, so no rung substitutes for another.

\section{MEASUREMENT RESULTS}
\label{sec:results}

\textbf{Task reward leaves most of it unexplained.} Across all $874$ measured morphologies,
competence correlates with task reward at Pearson $0.63$, $0.57$ and $0.32$ on Walker, Carrier and
DownStepper, so task reward accounts for $39\%$, $33\%$ and $10\%$ of the variance. Within the middle
decile of task reward ($32$, $28$ and $28$ designs), task reward spans $0.34$, $0.62$ and $0.45$
while competence spans $2.76$, $1.52$ and $2.48$ in the same units, so a co-design report would
place these morphologies close together.
Fig.~\ref{fig:problem} shows the extreme case, two bodies $0.2\%$ apart in task reward and
$2.8\times$ apart in what survives controller restriction.

\begin{figure*}[t]
  \centering
  \includegraphics[width=\textwidth]{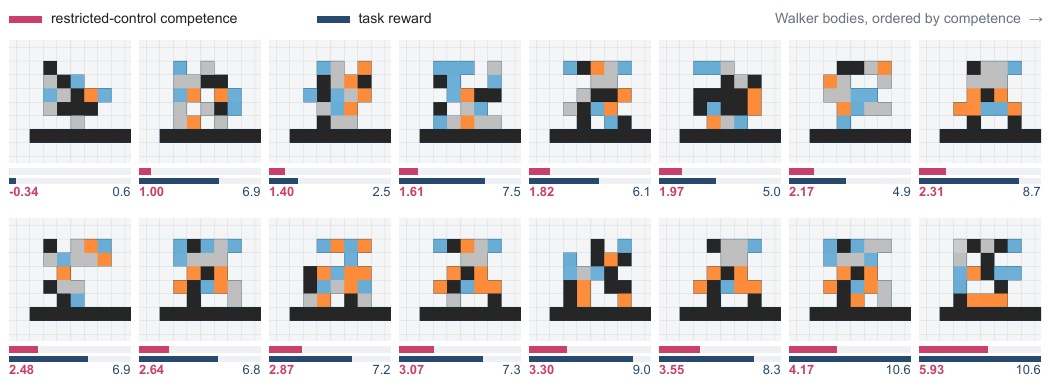}
  \caption{Sixteen co-designed Walker bodies, sampled evenly across the range of
  restricted-control competence and ordered by it; voxel colors as in Fig.~\ref{fig:tasks}. Under
  each body, the upper bar and bold number give its competence, the lower bar and plain number its
  task reward, on one shared scale. The ordering is not visible in the structure, which is why the
  score is measured.}
  \label{fig:gallery}
\end{figure*}

\textbf{Nor does the morphology's geometry predict it.} If the quantity were readable off the body,
measuring it would be unnecessary. We tested that with nine geometric features (actuator centroid,
spread and height, contact width and gaps, compactness, aspect ratio, and mass distribution),
cross-validated with folds grouped by evolutionary run so that a model cannot succeed by
recognizing a lineage it has seen. Ridge regression reaches $R^2 = -0.13$, $-0.08$ and $-0.03$ and
random forests $-0.14$, $-0.22$ and $+0.07$, where a negative value means the model does worse
than predicting the mean; adding task reward as a tenth feature gives $+0.00$, $+0.14$ and
$-0.05$. Fig.~\ref{fig:gallery} shows the same difficulty by eye. A property that neither the
reported performance number nor the visible geometry predicts has to be measured.

\textbf{Selecting on one declared family transfers only partly to another.} The arms with the score
in their objective were selected on the \emph{ID-phased} family, before the invariance problem was
fixed, so they answer a question we can now ask precisely. At matched task reward they raise the
ID-phased score they selected on by $+0.49 \pm 0.03$ on Walker and $+0.40 \pm 0.11$ on DownStepper
(mean $\pm$ s.d.\ across runs), while under the position-phased family the same designs retain
$+0.22 \pm 0.32$ and $+0.15 \pm 0.21$: roughly half the gain, with a spread an order of magnitude
wider. On Walker one of the three runs goes the wrong way ($-0.15$ against $+0.41$ and $+0.38$),
and on Carrier the direction reverses between families (ID-phased $-0.04$, position-phased
$+0.12$), though neither Carrier estimate is distinguishable from zero.

\textbf{Downstream correlates do not follow.} A high score suggests a body that is easier to build,
so we compared selection for the score against selection for reward on four such correlates, and
none held: tolerance to observation noise, actuator dropout and control latency; tolerance to an
imperfect driver (gain error $p = 0.12$, action noise $p = 0.54$, latency $p = 0.74$, clock error
$p = 0.95$); commanded actuation effort ($p = 0.16$); and controller size, where a linear policy
retains $73\%$ of full-capacity reward in the restricted-control-driven arm against $66\%$ in the
reward-only arm, at no capacity significantly. These design-level $p$ values treat designs as
independent and are therefore anti-conservative. The non-comparative part of
the premise does hold: replayed through a crude driver ($\pm 30\%$ per-actuator gain error, action
noise, four steps of latency, or $\pm 20\%$ clock error), each design's own restricted gait retains
$95$--$99\%$ of the score with the ordering intact ($\rho \ge 0.98$).

\section{DESIGN EXPERIMENTS}
\label{sec:design}

\subsection{Steering the search}

For each paired run we take the reward-only arm's
best task reward, keep every design in either arm within $5\%$ of it, and compare the two arms'
best competence inside that band. Both arms face the same absolute threshold, so an arm that
reached lower task reward gets no easier bar, and an arm with nothing in the band has failed to
reach competitive task performance, which counts as a loss rather than as missing data. (An
earlier version used hypervolume and could not be interpreted: its gap correlated $+0.83$ with the
gap in task reward each run happened to reach.) We ran NSGA-II~\cite{deb2002fast} on the pair
(task reward, C2 competence) against a reward-only arm on DownStepper, for five paired runs
matched in population ($24$), generations ($12$), training budget ($600$k steps) and a shared
generation-$0$ population whose best rewards agree exactly, so any divergence is attributable to
the selection rule. The bar, fixed in advance, was improvement in $4$ of $5$ runs with a positive
paired median, plus task reward non-inferior at a $5\%$ margin. Search selected on a score measured
at one optimizer restart, so every number below is a reevaluation at three fresh restarts.

\begin{table}[t]
\centering
\caption{Primary endpoint of the design experiment: the best reevaluated C2 competence among
designs within $5\%$ of the reward-only arm's best task reward (Thr.: that threshold; $n$: designs
in the band; MO: multi-objective arm). Two runs put nothing in the band.}
\label{tab:e4}
\footnotesize
\setlength{\tabcolsep}{4pt}
\begin{tabular}{lrrrrrr}
\hline
Seed & Thr. & MO & $n$ & Reward & $n$ & $\Delta$ \\
\hline
$0$ & $8.63$ & $5.33$ & $14$ & $4.55$ & $23$ & $+0.78$ \\
$1$ & $8.62$ & ---    & $0$  & $3.99$ & $3$  & empty \\
$2$ & $8.63$ & $3.97$ & $11$ & $3.40$ & $2$  & $+0.57$ \\
$3$ & $8.61$ & ---    & $0$  & $4.19$ & $7$  & empty \\
$4$ & $8.64$ & $2.78$ & $1$  & $2.07$ & $1$  & $+0.72$ \\
\hline
\end{tabular}
\end{table}

The multi-objective arm improved in $3$ of $5$ runs (Table~\ref{tab:e4}), against a bar of $4$,
with a paired median of $+0.72$ where both arms are defined. Task reward was not non-inferior: the
arms' ceilings agree to within $0.5\%$ in three runs but differ by $23\%$ and $31\%$ in the other
two ($t_4 = -0.88$, $p = 0.79$, one-sided). Neither pre-registered endpoint is met. One of the
three wins is also a sampling artifact, because the endpoint compares the maximum competence
inside the band and the maximum of $n$ draws grows with $n$: in run $2$ the multi-objective arm
placed $11$ designs in the band against the reward-only arm's $2$, and subsampling the larger arm
to the smaller arm's band size ($20{,}000$ draws) leaves runs $0$ and $4$ unaffected while run $2$
falls to a coin flip ($48.7\%$ of matched draws). The result is two clean wins, one sampling
artifact and two losses.

The two losses are a hitting time. Both are empty bands, so the endpoint never compared competence
at all: the task's landscape has a plateau near $6$--$7$ and a ceiling near $9.06$, and every run
that reached the ceiling, in either arm, crossed in a single generation and gained between $+2.04$
and $+2.31$. Reaching the band is therefore a discrete hitting event rather than a climb.
Excluding run $0$, whose generation-$0$ population already held a competitive design, the hitting
generations are $5$, $6$, $7$, $11$, $12$ and $12$, half of them in the final two generations, so
these runs are right-censored: one generation earlier the two arms would be tied at $2$ apiece, and
the observed $2$ of $4$ against $4$ of $4$ does not separate the two rules (descriptive one-sided
Fisher exact test, $p = 0.21$). Where the multi-objective arm made the jump, the second objective
cost no task performance, the paired ceilings agreeing to within $0.05$.

Two robustness checks, both specified before any of the data existed, bound the result. Retraining
the archive at twice the PPO budget ($1.2$M steps) moves arm ceilings by at most $0.03$, leaves the
count unchanged at $3$ of $5$ and the two empty bands empty, and leaves the multi-objective arm's
best archived design at only $77\%$ and $69\%$ of its comparator, far from the $95\%$ threshold:
the binding constraint is the evolutionary budget, not the training budget. The second check
measures competence with the held-out family C3, which no search selected on. The two families rank
the archive consistently ($\rho = +0.90$), and the agreement survives restricting C3 so that it
cannot reduce to C2, but the arm-level advantage does \emph{not} transfer cleanly: its sign agrees
in all three runs where the comparison exists while the magnitudes rearrange, the largest surviving
win (run $0$, $+0.78$ under C2) falling to $+0.07$ under C3.

\begin{table}[t]
\centering
\caption{Screening inside a reward-only run's top $5\%$ task-reward band. $\Delta$ is the
competence of the most competent band design minus that of the highest-reward one, under the
family used to choose (C2), the held-out family (C3), and C3 restricted so that it cannot reduce
to C2 (C3$'$). Cost is the task reward given up. D: DownStepper, W: Walker.
$^\dagger$Reward-only arm of the design experiment.}
\label{tab:screen}
\footnotesize
\setlength{\tabcolsep}{3.5pt}
\begin{tabular}{lrrrrr}
\hline
Run & $n$ & $\Delta$C2 & $\Delta$C3 & $\Delta$C3$'$ & Cost \\
\hline
D, s0$^\dagger$ & $23$ & $+2.62$ & $+2.23$ & $+1.52$ & $-0.015$ \\
D, s1$^\dagger$ & $3$  & $+0.45$ & $+0.65$ & $-0.11$ & $-0.002$ \\
D, s2$^\dagger$ & $2$  & $+1.19$ & $+1.01$ & $+1.43$ & $-0.048$ \\
D, s3$^\dagger$ & $7$  & $+0.70$ & $+1.38$ & $+1.72$ & $-0.008$ \\
D, s1    & $3$  & $+0.85$ & $+1.36$ & $+1.20$ & $-0.095$ \\
W, s2    & $12$ & $+2.48$ & $+1.72$ & $+0.67$ & $-0.056$ \\
W, s0    & $13$ & $+2.78$ & $+3.26$ & $+3.42$ & $-0.013$ \\
W, s1    & $8$  & $+1.40$ & $-0.25$ & $-0.15$ & $-0.002$ \\
\hline
\end{tabular}
\end{table}

\subsection{Screening after the search}

A cheaper use puts the measurement after the search: run an ordinary reward-only co-design search,
keep the designs within $5\%$ of its best task reward, and choose the most competent of them
instead of the highest-reward one, at the cost of one restricted-control pass over an existing
archive. We designed this analysis after the primary endpoints had been evaluated, so it is
exploratory, and we report every run. Screening needs a task-reward band holding more than one
design, which held in $8$ of $14$ reward-only runs: four of the design experiment's five, and four
of nine smaller population-$12$ runs, three on Walker and one on DownStepper. No Carrier run placed
more than one design in its band, so screening could not be evaluated there.

Where screening applies it always picks a different body (Table~\ref{tab:screen}), at a task-reward
cost of at most $0.10$ (median $0.01$) inside a band that allows about $0.5$. The chosen body
scores higher under C2 by construction, so the test is a family that played no part in the choice:
under C3 it is more competent in $7$ of $8$ runs (median $+1.37$), and under C3$'$, which cannot
reduce to C2, in $6$ of $8$ (median $+1.32$). One Walker run loses under both, so its advantage is
specific to the declared family. The count alone is suggestive (one-sided sign test, $p = 0.14$);
the stronger evidence is that the restriction leaves the median gain unchanged, which argues
against the families' overlap as its source. One operational requirement comes with the
recommendation: individual task rewards are not stable at our training budget, one chosen design
falling from $9.05$ to $2.02$ when retrained at $1.2$M steps, so a screened design should be
retrained before it is built.

The pre-registered claim, that selecting on competence during search finds morphologies a
reward-only search misses, is therefore not established at this scale. Three narrower results
stand: where the search reached the ceiling, selecting on the second axis cost no task
performance; of the $9$ non-dominated points in the union of the two arms, $7$ come from the
multi-objective arm; and where the arm reached competitive task performance it produced the more
competent morphology, by $+0.78$ and $+0.72$ under C2 after the band-size control. The variance
that limited the experiment is a hitting time on a discrete discovery: the remedy is to run the
same design until that distribution is no longer truncated.

\section{LIMITATIONS AND CONCLUSION}
\label{sec:limitations}

\textbf{Limitations.} The score inherits the choices that define the declared family: a fixed
amplitude, a fixed rest length, a sinusoidal waveform and phase set by position. A best-of-$Q$
grows with $Q$, so the reported value is budget-conditioned rather than a ceiling, and no family is
privileged: two defensible families rank the same morphologies at $0.50$--$0.63$, two
position-phased variants agree at only $0.41$ on Walker, and a third positional family agrees with
the declared one at $0.87$--$0.92$, so claims about a single body hold per declared family. The
simulator reset is deterministic and its physics fixed: perturbing the initial state retains
$97\%$ of the score and an imperfect actuator model $95$--$99\%$, but friction, stiffness, gravity
and mass are compiled in, with no domain randomization and no physical prototype, so any
buildability reading rests on driver-side evidence alone. All controllers use one
reinforcement-learning algorithm at a fixed $600{,}000$-step budget that is not at a plateau:
retraining eight Walker designs at $300{,}000$, $600{,}000$ and $1{,}200{,}000$ steps raises median
task reward from $4.36$ to $9.26$. Competence is \emph{bit-identical} at every budget because the
restricted controller is not trained, but the retention ratio's median falls from $0.152$ to
$0.141$, so the ratios we quote are inflated by about $8\%$ and are upper bounds, the ordering
surviving (Spearman $0.976$). Finally, the study uses one search method, one $5\times5$ grid, three
locomotion tasks and two to five runs per task and arm; a run is the independent unit, so that is a
small sample and the binding constraint on the design experiment, the screening result rests on
eight runs on two tasks and was designed after the pre-registered endpoints had been evaluated, and
nothing has been checked on hardware.

\textbf{Conclusion.} Co-design reports one number, and that number cannot distinguish a morphology
whose competence survives the removal of its controller from one whose competence does not. We
measure the difference by restricting the controller to a declared low-complexity family and
reporting the family, task, environment and budget with the score. On three EvoGym tasks the
quantity is reliable, converged and discriminative, largely unexplained by task reward and not
predicted by geometry, and it depends on the declared family strongly enough that two defensible
families reverse individual reward-matched pairs. That last finding is the one we would change the
field's practice on: a restricted-control number is interpretable only with its controller family
attached, so the useful object is a profile across families rather than a scalar. In design, the
measurement is already usable after the search: choosing among the bodies a reward-only run returns
changed the chosen body at negligible task-reward cost in every run offering a choice, and that
body usually stayed more competent under a family never used to choose it. Steering the search fell
short of a pre-registered bar at a $12$-generation budget, which the trajectories trace to when the
search reaches competitive bodies. The axis exists, the number the field reports does not contain
it, and measuring it is inexpensive and reproducible once the family is declared.

\section*{ACKNOWLEDGMENT}

The analysis and figure-generation code behind Secs.~\ref{sec:validity}--\ref{sec:design} and
Figs.~\ref{fig:problem}--\ref{fig:gallery}, and drafts of the manuscript text, were produced with
a generative AI system, Anthropic's Claude, through the Claude Code interface, under the authors'
direction. The authors specified and ran every experiment, verified every
reported number, and take responsibility for the content.



\begin{thebibliography}{99}

\bibitem{sims1994evolving} K. Sims, ``Evolving virtual creatures,'' in \emph{Proc. 21st Annu.
Conf. Comput. Graph. Interact. Tech. (SIGGRAPH)}, 1994, pp. 15--22.

\bibitem{lipson2000automatic} H. Lipson and J. B. Pollack, ``Automatic design and manufacture
  of robotic lifeforms,'' \emph{Nature}, vol. 406, pp. 974--978, 2000.

\bibitem{bhatia2021evolution} J. Bhatia, H. Jackson, Y. Tian, J. Xu, and W. Matusik, ``Evolution
Gym: A large-scale benchmark for evolving soft robots,'' in \emph{Adv. Neural Inf. Process. Syst.
(NeurIPS)}, vol. 34, 2021, pp. 2201--2214.

\bibitem{zhao2020robogrammar} A. Zhao, J. Xu, M. Konakovi\'{c}-Lukovi\'{c}, J. Hughes, A.
Spielberg, D. Rus, and W. Matusik, ``RoboGrammar: Graph grammar for terrain-optimized robot
design,'' \emph{ACM Trans. Graph.}, vol. 39, no. 6, art. 188, 2020.

\bibitem{hu2023glso} J. Hu, J. Whitman, and H. Choset, ``GLSO: Grammar-guided latent space
optimization for sample-efficient robot design automation,'' in \emph{Proc. Conf. Robot Learn.
(CoRL)}, PMLR vol. 205, 2023, pp. 1321--1331.

\bibitem{wang2019neural} T. Wang, Y. Zhou, S. Fidler, and J. Ba, ``Neural graph evolution: Towards
efficient automatic robot design,'' in \emph{Int. Conf. Learn. Represent. (ICLR)}, 2019.

\bibitem{yuan2022transform2act} Y. Yuan, Y. Song, Z. Luo, W. Sun, and K. Kitani,
``Transform2Act: Learning a transform-and-control policy for efficient agent design,'' in
\emph{Int. Conf. Learn. Represent. (ICLR)}, 2022.

\bibitem{gupta2021embodied} A. Gupta, S. Savarese, S. Ganguli, and L. Fei-Fei, ``Embodied
  intelligence via learning and evolution,'' \emph{Nature Communications}, vol. 12, art. 5721,
  2021.

\bibitem{lu2025bodygen} H. Lu, Z. Wu, J. Xing, J. Li, R. Li, Z. Li, and Y. Shi, ``BodyGen:
Advancing towards efficient embodiment co-design,'' in \emph{Int. Conf. Learn. Represent.
(ICLR)}, 2025.

\bibitem{wang2023diffusebot} T.-H. Wang, J. Zheng, P. Ma, Y. Du, B. Kim, A. Spielberg, J.
Tenenbaum, C. Gan, and D. Rus, ``DiffuseBot: Breeding soft robots with physics-augmented
generative diffusion models,'' in \emph{Adv. Neural Inf. Process. Syst. (NeurIPS)}, vol. 36,
2023, pp. 44398--44423.

\bibitem{liu2026morphology} S. Liu, J. Yan, H. Wang, and Y. Jin, ``Morphology evolution for
embodied robot design with a classifier-guided diffusion model,'' \emph{IEEE Trans. Evol.
Comput.}, vol. 30, no. 3, pp. 1039--1053, 2026.

\bibitem{chen2026natural} C. Chen, X. Xu, X. Wang, J. Wang, and X. Huang, ``Natural selection via
foundation models for soft robot evolution,'' in \emph{Proc. IEEE Int. Conf. Soft Robot.
(RoboSoft)}, 2026, pp. 422--429.

\bibitem{pfeifer2006body} R. Pfeifer and J. Bongard, \emph{How the Body Shapes the Way We Think: A
New View of Intelligence}. Cambridge, MA: MIT Press, 2006.

\bibitem{zahedi2013quantifying} K. Zahedi and N. Ay, ``Quantifying morphological computation,''
\emph{Entropy}, vol. 15, no. 5, pp. 1887--1915, 2013.

\bibitem{ghazizahedi2015quantifying} K. Ghazi-Zahedi and J. Rauh, ``Quantifying morphological
computation based on an information decomposition of the sensorimotor loop,'' in \emph{Proc. Eur.
Conf. Artif. Life (ECAL)}, 2015, pp. 70--77.

\bibitem{hauser2011towards} H. Hauser, A. J. Ijspeert, R. M. F\"{u}chslin, R. Pfeifer, and W.
Maass, ``Towards a theoretical foundation for morphological computation with compliant bodies,''
\emph{Biol. Cybern.}, vol. 105, no. 5--6, pp. 355--370, 2011.

\bibitem{haeufle2020morphological} D. F. B. Haeufle, K. Stollenmaier, I. Heinrich, S. Schmitt, and
K. Ghazi-Zahedi, ``Morphological computation increases from lower- to higher-level of biological
motor control hierarchy,'' \emph{Front. Robot. AI}, vol. 7, art. 511265, 2020.

\bibitem{ijspeert2008central} A. J. Ijspeert, ``Central pattern generators for locomotion
  control in animals and robots: A review,'' \emph{Neural Networks}, vol. 21, no. 4, pp.
  642--653, 2008.

\bibitem{xie2025morphology} Y. Xie, K.-f. Chu, X. Wang, and F. Iida, ``The morphology-control
trade-off: Insights into soft robotic efficiency,'' arXiv:2503.16127, 2025.

\bibitem{henderson2018deep} P. Henderson, R. Islam, P. Bachman, J. Pineau, D. Precup, and
  D. Meger, ``Deep reinforcement learning that matters,'' in \emph{Proc. AAAI Conf. Artificial
  Intelligence}, vol. 32, no. 1, 2018.

\bibitem{agarwal2021precipice} R. Agarwal, M. Schwarzer, P. S. Castro, A. Courville, and
  M. G. Bellemare, ``Deep reinforcement learning at the edge of the statistical precipice,''
  in \emph{Advances in Neural Information Processing Systems (NeurIPS)}, 2021.

\bibitem{deb2000efficient} K. Deb, ``An efficient constraint handling method for genetic
algorithms,'' \emph{Comput. Methods Appl. Mech. Eng.}, vol. 186, no. 2--4, pp. 311--338, 2000.

\bibitem{schulman2017proximal} J. Schulman, F. Wolski, P. Dhariwal, A. Radford, and O. Klimov,
``Proximal policy optimization algorithms,'' arXiv:1707.06347, 2017.

\bibitem{raffin2021stable} A. Raffin, A. Hill, A. Gleave, A. Kanervisto, M. Ernestus, and N.
Dormann, ``Stable-Baselines3: Reliable reinforcement learning implementations,'' \emph{J. Mach.
Learn. Res.}, vol. 22, no. 268, pp. 1--8, 2021.

\bibitem{holm1979simple} S. Holm, ``A simple sequentially rejective multiple test procedure,''
\emph{Scand. J. Statist.}, vol. 6, no. 2, pp. 65--70, 1979.

\bibitem{shrout1979intraclass} P. E. Shrout and J. L. Fleiss, ``Intraclass correlations: Uses in
assessing rater reliability,'' \emph{Psychol. Bull.}, vol. 86, no. 2, pp. 420--428, 1979.

\bibitem{deb2002fast} K. Deb, A. Pratap, S. Agarwal, and T. Meyarivan, ``A fast and elitist
multiobjective genetic algorithm: NSGA-II,'' \emph{IEEE Trans. Evol. Comput.}, vol. 6, no. 2,
pp. 182--197, 2002.

\end{thebibliography}
\end{document}